\pdfoutput=1
\documentclass[11pt]{article}
\usepackage[preprint]{acl}
\usepackage{times}
\usepackage{latexsym}
\usepackage[T1]{fontenc}
\usepackage[utf8]{inputenc}
\usepackage{microtype}
\usepackage{graphicx}
\usepackage{booktabs}
\usepackage{amsmath,amssymb}
\usepackage{multirow}
\usepackage{array}
\usepackage[table]{xcolor}
\usepackage{pifont}
\usepackage{caption}
\newcommand{\cmark}{\ding{51}}
\newcommand{\xmark}{\ding{55}}
\definecolor{khead}{HTML}{ECE9E1}
\definecolor{kblue}{HTML}{2A78D6}
\definecolor{korange}{HTML}{EB6834}
\definecolor{kred}{HTML}{C53030}
\definecolor{kgray}{HTML}{9A9893}
\definecolor{ksim}{HTML}{FDEEE7}
\definecolor{ktrue}{HTML}{E3EDFA}
\newcommand{\ndec}{1{,}590}
\newcommand{\optuniq}{96\%}

\makeatletter
\AtBeginDocument{\g@addto@macro\@maketitle{%
  \begin{center}
  \includegraphics[width=\textwidth]{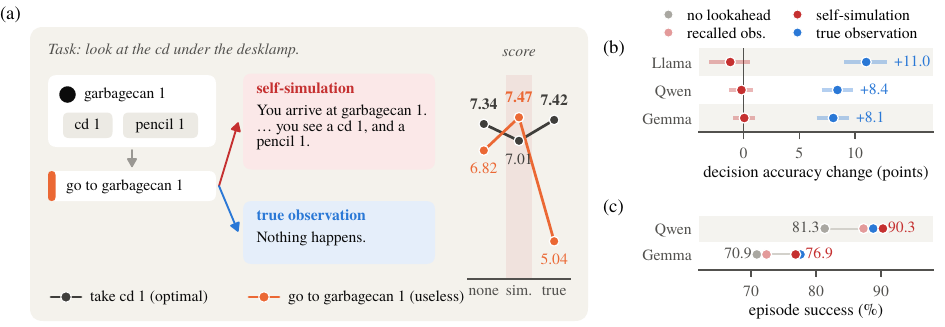}
  \captionof{figure}{Self-simulation tells an LLM agent what it already knows. (a) A simulation narrates the hoped-for outcome and flips the choice. (b) At one decision, true observations help, self-simulations do not. (c) In episodes, it matches true observations.}
  \label{fig:teaser}
  \end{center}
  \vspace{0.05in}}}
\makeatother

\title{Self-Simulation as Knowledge Restatement in LLM Agents}
\author{
  Mu Chen$^{1,*}$ \qquad Zhimei Chen$^{2}$ \\
  $^{1}$Mohamed bin Zayed University of Artificial Intelligence \qquad $^{2}$Independent Researcher \\
  \texttt{mu.chen@mbzuai.ac.ae} \\
  {\small $^{*}$Corresponding author: Mu Chen}
}

\hypersetup{pdftitle={Self-Simulation as Knowledge Restatement in LLM Agents},pdfauthor={Mu Chen, Zhimei Chen}}
\begin{document}
\maketitle
\begin{abstract}
Large language model (LLM) agents increasingly imagine the outcome of an action before acting. Top-leading solutions let the LLM simulate the next observation of each candidate action and choose among the imagined futures, crediting the gains to its knowledge of the environment. However, we find that self-simulation mostly tells an agent what it already knows, and that its value lies in restating the agent's own beliefs about each candidate at the point of choice. In ALFWorld, with planner-optimal actions as ground truth, true next observations raise the decision accuracy of three open LLMs by 8 to 11 points, whereas self-simulations change it by at most 1.2, since executability, the outcome that decides a choice, is simulated wrongly or already known. More importantly, even though two thirds of the targets it imagines are not there, self-simulation raises the episode success of Qwen-27B from 81.3\% to 90.3\%, as much as true observations (88.8\%), and a note that only recalls earlier observations recovers two thirds of this gain. We study one-step lookahead in a text household environment. Specifically, we build a placebo-controlled protocol that keeps the planner fixed, replaces only the note on each candidate, and scores every simulation against the environment.
\end{abstract}

\section{Introduction}

Recently, there has been a trend of model-based lookahead for large language model (LLM) agents, in which the LLM serves as its own world model: for every candidate action it predicts what the environment would show next, and a scorer ranks the candidates by their imagined outcomes. This design presumes that the LLM carries a usable model of its environment and that predicting the next state is what makes lookahead work. WebDreamer~\citep{gu2025your} supports this view by raising the success rate of a reactive GPT-4o web agent from 26.0\% to 37.0\% on Online-Mind2Web, WMA~\citep{chae2025web} feeds predicted transitions to a value function, and a growing family of agents learns, refines, or retrieves such simulations~\citep{fang2025webevolver,yu2026dynamind,liu2026imagine,wang2026memwm}. The presumption itself is rarely tested, because LLMs are unreliable simulators of text environments~\citep{wang2024can,li2025from} and the supporting ablations remove the simulation altogether~\citep{gu2025your,chae2025web}, which also removes the text that describes each candidate. Some results do not fit it: a prompted world model can underperform planning without any simulation, although the same agent chooses far better when it sees the true outcome of each candidate~\citep{chae2025web}.

However, one concern naturally arises that a self-simulation is written by the same model that scores it, and may therefore tell the scorer only what it already knows. Figure~\ref{fig:teaser}(a) tests this presumption on a single decision of Llama-8B in ALFWorld. The agent stands at garbagecan 1, on which it has just seen the cd 1 it must look at under the desklamp, and without lookahead it scores taking the cd highest (7.34), which is optimal. Its simulation of going to garbagecan 1, where it already stands, narrates an arrival that reveals the cd, the outcome it hopes for rather than the one the environment returns (``Nothing happens.''), so the score of this useless move rises from 6.82 to 7.47 and overtakes the optimal action; with the true observation it falls to 5.04. Over all \ndec{} decisions, true observations raise accuracy by 8.1 to 11.0 points for the three models, whereas self-simulation changes it by $-1.2$ to $+0.1$ (Figure~\ref{fig:teaser}(b), averages with 95\% intervals). More importantly, even though most of what it imagines is wrong, self-simulation helps over full episodes: it raises the success of Qwen-27B on 134 unseen games from 81.3\% to 90.3\%, as much as the true observation (88.8\%), and a note that only recalls what the agent saw earlier at each destination recovers 6.0 of these 9.0 points (Figure~\ref{fig:teaser}(c), where no lookahead denotes reranking of bare actions). The scorer thus reads its lookahead, yet a simulation it writes itself carries little that it lacks, which explains why prompted world models fall short of true outcomes. What a simulation contributes is a restatement of the agent's own beliefs about each candidate, so lookahead can work without predicting the next state correctly. Consequently, a more accurate world model alone does not close the gap, and the gains of lookahead measure how well the planner exposes the knowledge of the agent, not how well the LLM models its world.

\begin{figure*}[t]
  \centering
  \includegraphics[width=\textwidth]{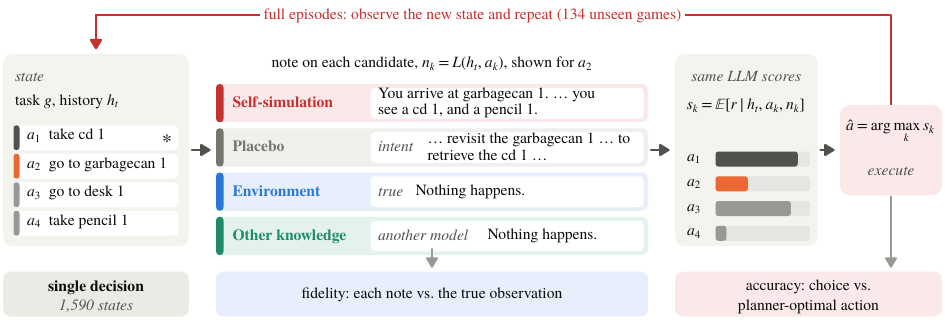}
  \caption{We keep a WebDreamer-style planner fixed and vary only the note the scorer reads with each candidate. Notes come from the LLM, placebos, the environment, or other knowledge; choices are evaluated at single decisions and over full episodes.}
  \label{fig:method}
\end{figure*}

In this work, we ask \emph{what an LLM agent learns from simulating its own actions}, and rethink lookahead in two aspects: what a self-simulation adds to a single choice, and what it adds over an episode. First, we hold a WebDreamer-style planner fixed and replace only the note on each candidate with placebo-controlled alternatives (Figure~\ref{fig:method}): length-matched generic text, a statement of intent, simulations shuffled across candidates, and the true next observation with its new content masked or its outcome flipped, so that each contrast changes one ingredient of the note while the planner stays the same. Second, we score every simulation against the environment and relate its fidelity to the choice of the agent and to what the agent knows without simulating, with optimal plan lengths from a PDDL planner as ground truth for \ndec{} decisions, so that every choice is graded exactly. Third, we run full episodes with the same notes and contrast the simulation with notes that only recall the history of the agent and with simulations shuffled across candidates, so that the gain in closed loop can be traced to the knowledge each note carries.

In summary, our work exhibits several compelling facets: First, it shows that the scorer reads its lookahead but gains nothing from its own, for Llama-3.1-8B, Qwen3.8-27B, and Gemma-4-31B alike. Shuffled or flipped true observations lower accuracy by 7.5 to 11.5 points, length-matched text changes it by at most 1.1, and a statement of intent adds at most 1.9. Second, it traces this null effect to executability, the only part of a simulation that predicts a correct choice ($\beta=4.7$ to $7.0$, $p<0.001$). Llama-8B reports 86\% of failed actions as executed, whereas Qwen-27B and Gemma-31B flag failures mostly where their scorers already rank the failed action last (83.6\% and 80.4\% of these decisions, against 62\% of the others). Knowledge from outside the scorer does help: simulations written by Qwen-27B raise the accuracy of Llama-8B by 5.0 points. Third, it explains the gain in closed loop. Self-simulation shortens the episodes of Qwen-27B from 17.3 to 13.7 steps and lowers revisits from 39\% to 32\%, the targets it imagines are there 32\% of the time against a base rate of 16\%, and shuffling the simulations across candidates costs 14.2 points, so what matters is that each note restates the beliefs of the agent about the candidate it accompanies.

\section{Related Work}

\paragraph{Model-Based Planning.}
A model of the environment lets an agent evaluate actions before it takes them, and LLMs make such a model available without task-specific training. Treating the language model as a world model and searching over its predictions dates back to RAP~\citep{hao2023reasoning}, and search over thoughts or actions with self-evaluation was popularized by Tree of Thoughts~\citep{yao2023tree} and LATS~\citep{zhou2024language}. For agents that act in environments, WebDreamer~\citep{gu2025your} simulates the outcome of each candidate action with the LLM and lets a scorer rank the simulations, and WMA~\citep{chae2025web} trains a dedicated world model whose predicted transitions feed a value function. A growing line of methods learns or refines such simulators~\citep{fang2025webevolver,yu2026dynamind,yu2026reinforcement,wang2026memwm,liu2026imagine,li2026discriminative} and couples them with lookahead in web and household environments, including ALFWorld~\citep{shridhar2021alfworld}. These works credit their gains to state prediction on the basis of ablations that remove the simulator altogether. We hold the planner fixed and replace only the content of the lookahead, which separates what the simulated state contributes from what an extra description of each candidate contributes.

\paragraph{LLMs as World Models.}
Language models are unreliable simulators of text environments~\citep{wang2024can,li2025from}, and they struggle to verify action preconditions in classical planning domains~\citep{valmeekam2023planning,kambhampati2024position}. With access to the true outcome of each candidate, LLMs choose the right action far more often~\citep{chae2025web}, yet a prompted world model can underperform planning without any simulation~\citep{chae2025web}, noisy foresight hurts while oracle foresight helps~\citep{zhang2026selfevolving}, and agents given an accurate simulator as a tool rarely use it well~\citep{qian2026current}. A ranker benefits most from predicted states that distinguish the candidates~\citep{li2026discriminative}. For learned world models, total prediction error correlates only weakly with planning quality, while error on decision-relevant variables correlates strongly~\citep{wang2026not}. We measure fidelity for each candidate action and relate it both to the choice of the scorer and to what the scorer already knows without simulation.

\paragraph{Placebo Controls.}
Placebo controls ask which part of an intermediate signal a model actually uses. Controls that keep the form of an intermediate signal while destroying its content have shown that random labels barely hurt in-context learning~\citep{min2022rethinking}, that invalid reasoning demonstrations retain most of the benefit of chain of thought~\citep{wang2023towards}, that filler tokens can stand in for reasoning~\citep{pfau2024lets}, and that the content of ReAct traces matters less than the similarity of exemplars~\citep{bhambri2025think}. Closest to our setting, two recent studies replace world-model training targets with mismatched or content-free signals and find that much of the gain survives~\citep{che2026learning,wang2026dark}. Those studies intervene during training; we intervene at inference, inside the lookahead loop, and find the opposite pattern for the scorer, which reads and follows the content it is given.

\paragraph{LLM Evaluators.}
The explanations of a model need not reflect what drives its decisions~\citep{turpin2023language,lanham2023measuring}, while LLM judges are sensitive to length, position, and verbosity~\citep{zheng2023judging,wang2024large,saito2023verbosity,dubois2024lengthcontrolled,ye2025justice}. Our length-matched generic text and shuffled conditions control for these effects, and the flipped condition directly tests whether the scorer conditions on the predicted outcome.

\section{Controlled Lookahead}
\label{sec:method}

We hold a WebDreamer-style planner fixed and vary only the text it reads about each candidate. We first formalize the planner and the shortcomings of current analyses (\S\ref{sec:problem}), then define the placebo-controlled notes (\S\ref{sec:conds}), the fidelity of a simulation (\S\ref{sec:fid}), and further variants of the note (\S\ref{sec:repair}).

\subsection{Preliminary}
\label{sec:problem}

At step $t$ an agent with task $g$ and interaction history $h_t$ holds a set of candidate actions $\mathcal{A}_t=\{a_1,\dots,a_K\}$. A lookahead planner writes a note $n_k=L(h_t,a_k)$ for every candidate, a scorer assigns $s_k=\mathbb{E}_{r\sim p_\theta(\cdot\mid h_t,a_k,n_k)}[r]$ with ratings $r\in\{1,\dots,9\}$, and the agent executes $\hat{a}=\arg\max_k s_k$. In WebDreamer~\citep{gu2025your}, $L$ asks the LLM to predict the observation that follows $a_k$, which we call \emph{self-simulation}; without lookahead, $n_k$ is empty and the scorer rates the bare action. Because ALFWorld is defined in PDDL, we can measure the quality of every choice exactly. Let $Q^*(s,a)$ be one plus the length of an optimal plan, computed with Fast Downward~\citep{helmert2006fast}, from the state reached by executing $a$ in $s$. A choice is correct when $\hat{a}\in\arg\min_k Q^*(s_t,a_k)$, and its regret is $Q^*(s_t,\hat{a})-\min_k Q^*(s_t,a_k)$ extra steps.

Current evidence for the role of the world model has two shortcomings. i) Ablations remove the simulation altogether, so the comparison changes the amount and the specificity of the text that the scorer reads at the same time as the predicted state. ii) World models are evaluated by downstream ranking accuracy or by aggregate prediction error, neither of which says which errors of a simulation change the choice of the agent.

\subsection{Placebo-Controlled Notes}
\label{sec:conds}

\begin{table}[t]
\centering
\small
\setlength{\tabcolsep}{3.2pt}
\begin{tabular}{lccccc}
\toprule
Condition & Text & Cand. & Exec. & Content & Source \\
\midrule
No lookahead & -- & -- & -- & -- & -- \\
Generic text & \cmark & \xmark & \xmark & \xmark & fixed \\
Intent only & \cmark & \cmark & \xmark & \xmark & LLM \\
Self-simulation & \cmark & \cmark & ? & ? & LLM \\
Shuffled simulation & \cmark & \xmark & ? & ? & LLM \\
True next state & \cmark & \cmark & \cmark & \cmark & env. \\
True dynamics & \cmark & \cmark & \cmark & \xmark & env. \\
Shuffled true state & \cmark & \xmark & \cmark & \cmark & env. \\
Flipped outcome & \cmark & \cmark & \xmark & \xmark & env. \\
\bottomrule
\end{tabular}
\caption{Lookahead conditions and what each note carries: any text, a description specific to the candidate, whether the action executes, and the newly observed content.}
\label{tab:conds}
\end{table}

Table~\ref{tab:conds} lists the conditions. They form contrasts that isolate one ingredient of the note at a time. \emph{Generic text} is a fixed neutral sentence truncated to the length of the self-simulation of the same candidate, which controls for the amount of text. \emph{Intent only} asks the same LLM to state what the action is meant to accomplish without predicting what the environment will show, which isolates the articulation of each candidate. \emph{Self-simulation} follows WebDreamer, with a prompt that lists example observations of the environment, including the failure message ``Nothing happens.'' \emph{Shuffled simulation} permutes the self-simulations among the candidates of a decision without fixed points, so the scorer reads a plausible outcome that belongs to another candidate.

Four conditions use the environment. \emph{True next state} is the real observation obtained by replaying the episode and executing the candidate. \emph{True dynamics} keeps this observation but replaces every list of newly visible objects with ``[contents not shown]'', so it reveals whether and how the action changes the state, without information that the agent could not have predicted. \emph{Shuffled true state} permutes the true observations among candidates, and \emph{flipped outcome} turns every executed action into ``Nothing happens.'' and every failed action into the message its success would produce. If the scorer reads the outcome, the last two conditions should fall below no lookahead; if it ignores the outcome, they should match no lookahead.

\subsection{Simulation Fidelity}
\label{sec:fid}

We score each self-simulation against the true observation on two variables. \emph{Executability} asks whether the simulation and the environment agree that the action takes effect: a simulation predicts failure when it reports that nothing happens or that the action cannot be performed. \emph{Content} is the F1 overlap between the object instances (``mug 1'') that the simulation and the environment newly reveal, excluding objects named in the action, and is defined for executable actions that reveal objects. To relate fidelity to choice, we regress the correctness of the self-simulation choice on the fraction of candidates whose executability is predicted correctly and on mean content F1, controlling for whether the choice without lookahead is correct and for the length of the simulations, with standard errors clustered by game.

\subsection{Note Variants}
\label{sec:repair}

Two repairs test whether fixing executability alone recovers the benefit of true outcomes. The \emph{oracle repair} keeps each self-simulation but corrects its executability with the environment: a simulation of a failed action becomes ``Nothing happens.'' and a simulation that wrongly predicts failure is replaced by the true dynamics. \emph{Precondition-checked simulation} requires no environment access: the world-model prompt first asks the LLM to state in one sentence whether the preconditions of the action hold (for example, that the agent stands at the receptacle it takes from, or holds the object it moves), and then to write the observation. Only the observation enters the note. A third variant, \emph{verdict only}, replaces the note with a single sentence stating whether the precondition-checked simulation predicts that the action executes.

Three further variants probe the source of the information. \emph{Cross-model simulation} gives each scorer the simulations that another model wrote for the same candidates, which tests whether a simulation helps once it carries knowledge the scorer lacks. \emph{Effect description} replaces the exact observation with a short description of what the action changes, closer to the wording of the WebDreamer prompt~\citep{gu2025your}. \emph{Recalled observation} predicts nothing: for a navigation or opening action it repeats the last observation the agent made at the destination, or states that the agent has not been there, and for other actions it states that no earlier observation applies.

\paragraph{Inference and Complexity.}
All conditions use one scorer call per candidate, as in WebDreamer, and the LLM-written notes add one generation per candidate. The precondition check adds about one sentence to each generation and no further calls. The environment conditions require privileged access and serve only as references for analysis.

\section{Experimental Setup}
\label{sec:setup}

\paragraph{Environment and Decision Points.}
We use the text version of ALFWorld~\citep{shridhar2021alfworld}, built on TextWorld~\citep{cote2018textworld}, whose six task families require finding, transforming (heating, cooling, cleaning), and placing household objects. Decision points come from all 274 validation games (140 seen and 134 unseen). For each game we roll out the planner policy twice, replacing its action with a random admissible one with probability 0.25 after the first step, so that the states include detours and recoverable mistakes, and we sample three steps from each rollout. Rollouts in which the planner did not return within 150\,s (18 of 548) are dropped, which leaves \ndec{} decision points from 267 games (Appendix~\ref{app:data}). Each has four candidates in random order: the planner action, two other admissible actions, and one action that was admissible elsewhere in the same episode but cannot be executed in the current state, such as taking an object from a receptacle the agent has left. Candidates of the last kind are common in practice, since LLM agents frequently propose actions whose preconditions do not hold. Exactly one candidate is optimal at \optuniq{} of decision points, so a uniform choice is correct about 26\% of the time.

\paragraph{Models and Scoring.}
We analyze Llama-3.1-8B-Instruct~\citep{grattafiori2024llama}, Qwen3.8-27B~\citep{yang2025qwen3} in non-thinking mode, and Gemma-4-31B-it~\citep{team2025gemma}, which we call Llama-8B, Qwen-27B, and Gemma-31B, and serve them with vLLM~\citep{kwon2023efficient}. Within each experiment one model plays every role, as in WebDreamer: it writes the simulation and the intent, and it scores. Notes are decoded greedily. The scorer rates each candidate on a 1 to 9 scale, and we take the expected rating under its next-token distribution over the nine digits, which equals the average over infinitely many sampled ratings and avoids ties. A listwise scorer that sees all four candidates with their notes and names the best one is reported in the appendix, together with the full prompts of every role.

\paragraph{Episodes.}
To test whether the decision-level effects reach task success, we run full episodes on the 134 unseen games with at most 40 steps. At every step the agent samples eight actions at temperature 1 from a prompt with two expert demonstrations from the training split, keeps the four most frequent distinct actions as candidates, writes a note for each according to the condition, and executes the candidate with the highest score. The true-state condition replays the episode to obtain the real next observation of every candidate. The reactive baseline executes the greedy action without lookahead, and reranking without lookahead scores the bare candidates. We run reactive, reranking, intent, recalled observation, self-simulation, and true-state notes for both larger models, and shuffled and precondition-checked simulations for Qwen-27B.

\paragraph{Statistics.}
All confidence intervals are 95\% percentile intervals from bootstrap resamples of games (2{,}000 for decisions and 4{,}000 for episodes), so that decision points from the same game move together. Differences between conditions are paired by decision, since all conditions score the same candidates, and by game in episodes.

\section{Self-Simulation at a Single Decision}
\label{sec:results}

\begin{table}[t]
\centering
\footnotesize
\setlength{\tabcolsep}{2.4pt}
\begin{tabular}{lrrr}
\toprule
\rowcolor{khead} Note & Llama-8B & Qwen-27B & Gemma-31B \\
\midrule
No lookahead & 67.7 & 85.3 & 85.3 \\
\midrule
Generic text & 66.5 {\scriptsize \textcolor{kgray}{-1.1}} & 85.2 {\scriptsize \textcolor{kgray}{-0.2}} & 85.2 {\scriptsize \textcolor{kgray}{-0.1}} \\
Intent only & 69.6 {\scriptsize \textcolor{kgray}{+1.9}} & 86.4 {\scriptsize \textcolor{kgray}{+1.0}} & 87.0 {\scriptsize \textcolor{kblue}{+1.8}} \\
Shuffled simulation & 62.5 {\scriptsize \textcolor{korange}{-5.2}} & 78.2 {\scriptsize \textcolor{korange}{-7.1}} & 77.0 {\scriptsize \textcolor{korange}{-8.2}} \\
\rowcolor{ksim} Self-simulation & 66.5 {\scriptsize \textcolor{kgray}{-1.2}} & 85.2 {\scriptsize \textcolor{kgray}{-0.2}} & 85.3 {\scriptsize \textcolor{kgray}{+0.1}} \\
\midrule
Flipped outcome & 57.9 {\scriptsize \textcolor{korange}{-9.7}} & 77.8 {\scriptsize \textcolor{korange}{-7.5}} & 75.5 {\scriptsize \textcolor{korange}{-9.8}} \\
Shuffled true state & 56.2 {\scriptsize \textcolor{korange}{-11.5}} & 74.5 {\scriptsize \textcolor{korange}{-10.8}} & 74.2 {\scriptsize \textcolor{korange}{-11.1}} \\
True dynamics & 74.8 {\scriptsize \textcolor{kblue}{+7.2}} & 88.0 {\scriptsize \textcolor{kblue}{+2.6}} & 87.8 {\scriptsize \textcolor{kblue}{+2.5}} \\
\rowcolor{ktrue} True next state & 78.7 {\scriptsize \textcolor{kblue}{+11.0}} & 93.8 {\scriptsize \textcolor{kblue}{+8.4}} & 93.3 {\scriptsize \textcolor{kblue}{+8.1}} \\
\bottomrule
\end{tabular}
\caption{Decision accuracy (\%) with the change against no lookahead; colored changes have 95\% game-bootstrap intervals that exclude zero; changes are computed before rounding. A uniform choice scores about 26\%.}
\label{tab:main}
\end{table}

\subsection{Sensitivity to Note Content}

The scorer conditions on what a note says about the outcome, which rules out the simplest explanation for a null result (Table~\ref{tab:main}). Notes that describe the wrong candidate lower accuracy well below no lookahead: shuffled simulations cost 5.2, 7.1, and 8.2 points for Llama-8B, Qwen-27B, and Gemma-31B, and shuffled true observations cost 10.8 to 11.5 points. Flipping the outcome of each true observation costs 7.5 to 9.8 points and raises the share of chosen actions that cannot be executed from 20.7\% to 30.1\%, 3.7\% to 12.4\%, and 2.6\% to 9.1\% (Table~\ref{tab:app_list}). In the other direction, the true next observation raises accuracy by 11.0, 8.4, and 8.1 points, halves the regret of the two larger models, and lowers it by 30\% for Llama-8B. Length alone does nothing, since the generic text of matched length changes accuracy by at most 1.1 points. The planner can therefore turn correct lookahead into better choices, so a null effect of self-simulation cannot come from an inattentive scorer.

\subsection{Effect of Self-Simulation}

Against this backdrop, self-simulation leaves accuracy unchanged for all three models ($-1.2$, $-0.2$, and $+0.1$ points, with 95\% intervals that include zero). A statement of intent, which describes each candidate without predicting an outcome, adds at most 1.9 points, so the articulation of the candidate is not what lookahead supplies at a single decision either. The self-simulation also moves the scorer the least: after centering within each decision, candidate scores with and without it correlate at 0.94 to 0.97, more closely than with the intent or the true observation (Table~\ref{tab:app_reg}). The listwise scorer, which compares all candidates and their notes side by side, shows the same pattern for Llama-8B and Qwen-27B (Table~\ref{tab:app_list}). An effect description in the style of WebDreamer helps only Llama-8B (2.7 points) and costs the larger models about 3 points (Table~\ref{tab:repair}).

\subsection{Simulation Fidelity Analysis}
\label{sec:knows}

\begin{figure*}[t]
  \centering
  \includegraphics[width=\textwidth]{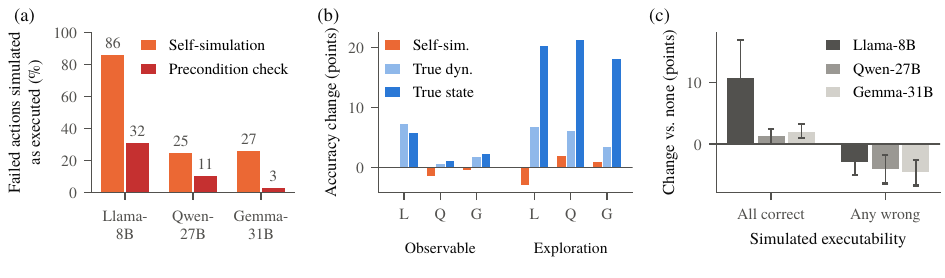}
  \caption{Executability decides what a self-simulation adds. (a) Failed actions simulated as executed. (b) Accuracy change at observable and exploration decisions. (c) Change from self-simulation when its executability is all right or partly wrong.}
  \label{fig:analysis}
\end{figure*}

A note that the scorer reads adds nothing when the scorer's own model writes it, because the content that decides a choice is executability, which this model either simulates wrongly or already uses. Llama-8B reports as executed 86\% of the actions that cannot be executed (Figure~\ref{fig:analysis}(a)), so its simulations mostly narrate the intended effect of each action; it reports every redundant move to its current location as an arrival, as in Figure~\ref{fig:teaser}(a), and 95\% of moves of an object it does not hold as successful (Table~\ref{tab:app_fail}). Qwen-27B and Gemma-31B are far more accurate, reporting 25\% and 27\% of failed actions as executed. All three are poor at content: the F1 between the objects that a simulation and the environment reveal is 0.16 to 0.18. In a logistic regression that controls for whether the choice without lookahead is correct, the share of candidates with correctly simulated executability predicts a correct choice with self-simulation for every model ($\beta=4.9$, $4.7$, and $7.0$, all $p<0.001$), whereas content F1 matters only weakly for Llama-8B ($\beta=0.9$, $p=0.006$) and not for the others ($p>0.8$). Figure~\ref{fig:analysis}(c) shows the same relation: simulations that get executability right for every candidate help Llama-8B by 10.7 points, and a single error turns the change negative for every model.

Accurate simulation still does not translate into a gain, because an accurate simulation says what its scorer already uses. Without any lookahead, the two larger models choose a non-executable action in only 3.7\% and 2.6\% of decisions, and their scorers rank the non-executable candidate lowest in 58.7\% and 61.9\% of decisions. Their simulations flag the failure in 83.6\% and 80.4\% of exactly these decisions but in only 62.3\% and 62.2\% of the others, so a simulation is mostly right where the scorer was already right. For Llama-8B neither signal is informative: its scores without lookahead separate executable from non-executable candidates at chance (AUROC 0.50), and its simulations barely better (0.57). A model that simulates its own actions thus presents its knowledge twice, once in the score and once in the simulation, which adds nothing when the knowledge is missing and little when it is present.

\subsection{Effect of External Knowledge}

\begin{table}[t]
\centering
\footnotesize
\setlength{\tabcolsep}{1.4pt}
\begin{tabular}{lrrr}
\toprule
\rowcolor{khead} Note & Llama-8B & Qwen-27B & Gemma-31B \\
\midrule
\rowcolor{ksim} Self-simulation & 66.5 {\scriptsize \textcolor{kgray}{-1.2}} & 85.2 {\scriptsize \textcolor{kgray}{-0.2}} & 85.3 {\scriptsize \textcolor{kgray}{+0.1}} \\
\midrule
Oracle exec. fix & 72.8 {\scriptsize \textcolor{kblue}{+5.2}} & 87.5 {\scriptsize \textcolor{kblue}{+2.1}} & 87.6 {\scriptsize \textcolor{kblue}{+2.3}} \\
Precondition check & 67.2 {\scriptsize \textcolor{kgray}{-0.4}} & 86.7 {\scriptsize \textcolor{kblue}{+1.3}} & 85.1 {\scriptsize \textcolor{kgray}{-0.2}} \\
Verdict only & 67.0 {\scriptsize \textcolor{kgray}{-0.6}} & 85.5 {\scriptsize \textcolor{kgray}{+0.1}} & 83.0 {\scriptsize \textcolor{korange}{-2.3}} \\
Effect description & 70.4 {\scriptsize \textcolor{kblue}{+2.7}} & 82.5 {\scriptsize \textcolor{korange}{-2.9}} & 82.3 {\scriptsize \textcolor{korange}{-3.0}} \\
Recalled observation & 64.3 {\scriptsize \textcolor{korange}{-3.4}} & 85.3 {\scriptsize \textcolor{kgray}{-0.1}} & 84.8 {\scriptsize \textcolor{kgray}{-0.5}} \\
\midrule
Sim. of Qwen-27B & 72.6 {\scriptsize \textcolor{kblue}{+5.0}} & -- & 84.3 {\scriptsize \textcolor{kgray}{-1.0}} \\
Sim. of Llama-8B & -- & 82.1 {\scriptsize \textcolor{korange}{-3.3}} & 82.8 {\scriptsize \textcolor{korange}{-2.5}} \\
\bottomrule
\end{tabular}
\caption{Variants of the note: decision accuracy (\%) with the change against no lookahead, colored as in Table~\ref{tab:main}. The last two rows give each scorer the self-simulations written by another model.}
\label{tab:repair}
\end{table}

If a self-simulation adds nothing because it repeats the knowledge of its own scorer, a simulation that carries knowledge the scorer lacks should help. Table~\ref{tab:repair} confirms this with simulations written by another model. Giving Llama-8B the simulations that Qwen-27B wrote for the same candidates raises its accuracy by 5.0 points, whereas its own simulations change it by $-1.2$, and the share of its choices that cannot be executed falls from 20.7\% to 12.5\%. The same simulations do not help Gemma-31B ($-1.0$), whose own knowledge of executability is as good, and the optimistic simulations of Llama-8B hurt both larger models ($-3.3$ and $-2.5$). Correcting only the executability of each self-simulation with the environment helps all three models (5.2, 2.1, and 2.3 points), most of the gain of the full true dynamics. Asking the model itself to check preconditions before it simulates does not reliably substitute for the environment. The check lowers the share of failed actions reported as executed to 32\%, 11\%, and 3\% (Figure~\ref{fig:analysis}(a)), but it gains 1.3 points only for Qwen-27B; Llama-8B now also reports a quarter of executable actions as failures, and the larger models gain little that they did not already use. Recalling the last observation the agent made at the destination of each candidate, a note built from the history alone, does not help either, and costs Llama-8B 3.4 points.

\subsection{Dynamics and Content}

The true next observation helps because it brings information the model cannot generate. Table~\ref{tab:main} splits its gain into dynamics and content. True dynamics with masked content recover 7.2 of the 11.0 points for Llama-8B, which lacks executability knowledge, but only 2.6 and 2.5 of the roughly 8 points for the two larger models, which already have it. The rest comes from content. Figure~\ref{fig:analysis}(b) separates the 572 exploration decisions, whose optimal action is to visit or open an unseen receptacle, from the 1{,}018 decisions in which the optimal action is determined by what the agent has observed (L, Q, and G denote the three models). At exploration decisions the true observation raises accuracy by 20.3, 21.3, and 18.1 points, whereas self-simulation changes it by $-3.0$ to $+2.1$; at observable decisions the larger models are already accurate without lookahead. Exploration decisions are therefore where lookahead could add the most, and there a self-simulation can contribute only the beliefs of the agent about where objects tend to be, which Section~\ref{sec:closed} shows to matter over an episode.

\section{Self-Simulation in Closed Loop}
\label{sec:closed}

\begin{table}[t]
\centering
\small
\setlength{\tabcolsep}{2.6pt}
\begin{tabular}{lrrrr}
\toprule
\rowcolor{khead} Note & Success & Steps & Revisit & Hit \\
\midrule
\multicolumn{5}{l}{\textit{Qwen-27B}} \\
Reactive & 74.6 & 17.4 & 51 & 17 \\
Reranking, no lookahead & 81.3 & 17.3 & 39 & 16 \\
\midrule
Intent only & 86.6 {\scriptsize\textcolor{kblue}{+5.2}} & 15.8 & 31 & 16 \\
Recalled observation & 87.3 {\scriptsize\textcolor{kblue}{+6.0}} & 14.7 & 29 & 18 \\
Shuffled simulation & 76.1 {\scriptsize\textcolor{kgray}{-5.2}} & 20.8 & 40 & 15 \\
\rowcolor{ksim} Self-simulation & 90.3 {\scriptsize\textcolor{kblue}{+9.0}} & 13.7 & 32 & 19 \\
Precondition-checked sim. & 89.6 {\scriptsize\textcolor{kblue}{+8.2}} & 13.7 & 33 & 19 \\
\rowcolor{ktrue} True next state & 88.8 {\scriptsize\textcolor{kblue}{+7.5}} & 12.7 & 35 & 25 \\
\midrule
\multicolumn{5}{l}{\textit{Gemma-31B}} \\
Reactive & 77.6 & 17.3 & 40 & 19 \\
Reranking, no lookahead & 70.9 & 19.2 & 40 & 18 \\
\midrule
Intent only & 73.9 {\scriptsize\textcolor{kgray}{+3.0}} & 18.4 & 37 & 18 \\
Recalled observation & 72.4 {\scriptsize\textcolor{kgray}{+1.5}} & 18.5 & 37 & 18 \\
\rowcolor{ksim} Self-simulation & 76.9 {\scriptsize\textcolor{kblue}{+6.0}} & 18.0 & 39 & 18 \\
\rowcolor{ktrue} True next state & 77.6 {\scriptsize\textcolor{kblue}{+6.7}} & 17.3 & 38 & 20 \\
\bottomrule
\end{tabular}
\caption{Episodes on the 134 unseen games: success (\%) with the change against reranking without lookahead, mean steps, share of navigation steps that revisit a receptacle (\%), and share of first visits that reveal the target object (\%).}
\label{tab:ep}
\end{table}

\begin{figure}[t]
  \centering
  \includegraphics[width=\columnwidth]{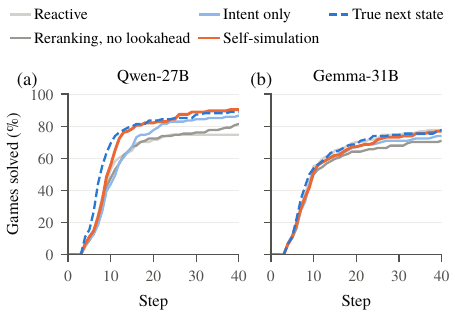}
  \caption{Over full episodes, self-simulation ends with as many solved games as true next observations and more than reranking without lookahead. Curves show, for (a) Qwen-27B and (b) Gemma-31B, the share of the 134 unseen games solved by each step.}
  \label{fig:ep}
\end{figure}

Over an episode, the agent proposes its own candidates and its history grows. Table~\ref{tab:ep} reports full episodes of the two larger models on the 134 unseen games; Llama-8B solved only 9 of 274 games within 16 steps in a pilot run, too few for a comparison. Reranking the agent's own proposals changes success from the reactive baseline by $+6.7$ points for Qwen-27B and by $-6.7$ points for Gemma-31B, whose scorer of bare actions prefers worse proposals than its own greedy choice. On top of reranking, self-simulation adds 9.0 and 6.0 points (95\% intervals 3.7 to 14.2 and 0.7 to 11.2), and the true next observation adds 7.5 and 6.7 points. In closed loop, unlike at a single decision, the simulation thus matches an oracle that sees the real outcome of every candidate. For Qwen-27B, self-simulation pulls ahead of reranking between steps 10 and 15 and stays ahead to the end (Figure~\ref{fig:ep}; 78.4\% against 67.2\% solved by step 15).

The gain does not come from predicting the world. For Qwen-27B, 24\% of the simulated moves to a receptacle that the agent has not yet visited place the target object there, and when the agent follows such a simulation, the target is actually there 32\% of the time. This is about twice the rate at which a first visit reveals the target without lookahead (16\%), so the imagined contents carry the agent's beliefs about where objects tend to be, but two thirds of them are wrong. The simulation also changes how the agent moves: it lowers the share of navigation steps that return to an already visited receptacle from 39\% to 32\% and shortens episodes from 17.3 to 13.7 steps, because the simulation of a revisit repeats what the agent saw there. Without lookahead, the agent often sweeps the same row of shelves twice (Appendix~\ref{app:episodes}).

Both effects restate knowledge the agent already has, and notes that contain only this knowledge recover most of the gain for Qwen-27B. The recalled observation, which predicts nothing, adds 6.0 points for Qwen-27B and lowers revisits to 29\%, and a statement of intent adds 5.2 points; for Gemma-31B the two notes add 1.5 and 3.0 points, short of significance. The simulation must, however, be about the candidate it is attached to: shuffling the simulations across candidates lowers success to 76.1\%, 14.2 points below self-simulation (95\% interval 7.5 to 20.9), lengthens episodes to 20.8 steps, and the imagined targets the agent follows are then there only 17\% of the time, close to the base rate. Checking preconditions before simulating changes little in closed loop (89.6\% against 90.3\%). At a single decision with a fixed set of candidates, the scorer already uses what it knows about them, so a restatement adds nothing; over an episode, a bare action string leaves the scorer to search a long history for what the candidate implies, and the restatement performs this search at the point of choice.

\section{Conclusion}

A self-simulation tells an LLM agent what it already knows. At a single decision, the scorer reads its notes, but a simulation written by its own model carries little that it lacks: a weaker model narrates the intended effect of actions that cannot be executed, and a stronger model flags failures mostly where it already avoids them. Over an episode, the same restatement brings the memory and beliefs of the agent to the point of choice, so self-simulation matches a true one-step oracle although most of its imagined contents are wrong. Lookahead should therefore be credited by the information it adds beyond what the policy already uses, which placebo and true-state references such as ours measure.

\section*{Limitations}

Although ALFWorld provides exact optimal plans and thus an exact grade for every choice, it is a single text environment with deterministic dynamics, and web environments with richer and noisier observations may shift the balance between executability and content. Our analysis covers three open models from 8B to 31B parameters with one-step lookahead and one prompt per role, and the episodes use the two larger models with a single sampling seed, so larger proprietary models, deeper search, and learned world models may behave differently. The decision points come from perturbed planner rollouts with exactly one non-executable candidate each, a design chosen for measurement that the episode experiments only partly relax. Extending the placebo-controlled protocol to web agents and to multi-step lookahead is a natural next step, since both change what a simulation can add beyond the policy.

\section*{Broader Impact and Ethical Considerations}

LLM agents increasingly plan by imagining the outcomes of their actions, and our findings suggest that much of the value of such lookahead lies in restating what the agent already knows at the point of choice. The same view may apply wherever a model imagines or critiques its own outputs, as in self-refinement and self-evaluation, and it implies that evaluations of world models for agents should report placebo and true-state references next to downstream success, and that agent designs can obtain part of the benefit from cheaper notes that recall the history. On the negative side, an agent that trusts its own optimistic simulations may take actions whose preconditions do not hold, which can cause unintended side effects in real systems, and notes of the kind we study could be crafted to steer the choices of an agent. Checking executability with the environment or a stronger model before acting mitigates the first risk, and auditing the notes an agent reads mitigates the second. This work uses publicly released models in a simulated household environment and involves no human subjects, crowdsourcing, or personal data.

\bibliography{refs}

\appendix

\section{Decision Points}
\label{app:data}

\begin{figure*}[t]
  \centering
  \includegraphics[width=\textwidth]{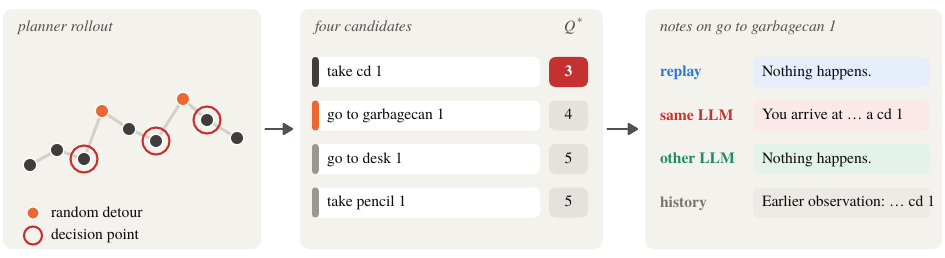}
  \caption{A decision point is sampled from a perturbed planner rollout, each candidate is graded by its cost-to-go $Q^{*}$ from a PDDL planner, and its notes come from replaying the game, the scoring LLM, another LLM, or the history.}
  \label{fig:app_pipeline}
\end{figure*}

Figure~\ref{fig:app_pipeline} shows how a decision point and its notes are built on the example of Figure~\ref{fig:teaser}. The \ndec{} decision points come from 267 of the 274 validation games (801 from seen and 789 from unseen games); the remaining seven games lost both rollouts to planner time-outs. By task family there are 354 pick-and-place, 339 clean, 267 cool, 231 heat, 213 pick-two, and 186 look-at decisions. Decisions occur at a median of step 3 (maximum 20), and the optimal remaining plan from the decision state has a median of 3 steps. Of the 1{,}657 optimal candidates (64 decisions have tied optima), 853 are navigation actions, 314 take actions, 241 move actions, and 249 open, use, cool, clean, or heat actions. A non-optimal admissible candidate costs 1.51 extra steps on average, and the non-executable candidate costs one step, since it leaves the state unchanged.

The cost-to-go $Q^*$ is computed by replaying the action prefix in a fresh copy of the game, executing the candidate, and calling the Fast Downward planner bundled with TextWorld on the resulting PDDL state. The planner knows the location of every object, so $Q^*$ is optimal under full information. We call a decision an exploration decision when its optimal candidate goes to or opens a receptacle that has not been mentioned in the history; 572 decisions are of this kind.

The four conditions that use the environment are constructed as follows. The true next observation is the feedback string returned by the environment after the candidate. True dynamics replaces every span matching ``you see \dots.'' with ``you see [contents not shown].'', which removes lists of objects and the word ``nothing'' while keeping arrivals, door states, pick-ups, and failures. The flipped outcome replaces any successful feedback with ``Nothing happens.'' and replaces ``Nothing happens.'' with a template of the successful feedback for the action (for example ``You pick up the X from the Y.''). The generic text repeats the sentence ``This is one of several candidate actions that the agent could take at the current point of the episode, and it is listed here so that it can be evaluated together with the other options that the planner is considering at this step of the interaction with the environment.'' and truncates it to the word count of the self-simulation of the same candidate. Shuffled conditions apply a random derangement of the candidates within each decision.

\section{Additional Results}
\label{app:results}

\begin{table*}[t]
\centering
\small
\setlength{\tabcolsep}{3pt}
\begin{tabular}{lrrrrrrrrr}
\toprule
 & \multicolumn{3}{c}{Llama-8B} & \multicolumn{3}{c}{Qwen-27B} & \multicolumn{3}{c}{Gemma-31B} \\
\cmidrule(lr){2-4} \cmidrule(lr){5-7} \cmidrule(lr){8-10}
Note & List acc. & Regret & Non-exec. & List acc. & Regret & Non-exec. & List acc. & Regret & Non-exec. \\
\midrule
No lookahead & 70.3 & 0.34 & 20.7 & 87.4 & 0.17 & 3.7 & 63.6 & 0.17 & 2.6 \\
Generic text & 66.7 & 0.36 & 20.3 & 86.8 & 0.17 & 4.2 & 70.3 & 0.17 & 2.7 \\
Intent only & 68.1 & 0.34 & 15.8 & 86.0 & 0.16 & 3.1 & 79.2 & 0.16 & 2.3 \\
Shuffled simulation & 56.8 & 0.41 & 19.1 & 66.5 & 0.25 & 6.0 & 36.8 & 0.27 & 4.3 \\
Self-simulation & 67.5 & 0.35 & 20.8 & 86.8 & 0.17 & 3.9 & 69.2 & 0.18 & 2.8 \\
Flipped outcome & 44.4 & 0.45 & 30.1 & 80.8 & 0.24 & 12.4 & 50.0 & 0.31 & 9.1 \\
Shuffled true state & 54.0 & 0.48 & 24.3 & 64.1 & 0.28 & 6.2 & 32.5 & 0.31 & 5.2 \\
True dynamics & 72.7 & 0.27 & 8.7 & 88.9 & 0.14 & 0.9 & 67.1 & 0.15 & 0.9 \\
True next state & 79.7 & 0.24 & 10.3 & 95.2 & 0.08 & 0.7 & 74.5 & 0.09 & 0.7 \\
Oracle executability repair & 72.7 & 0.29 & 10.1 & 87.9 & 0.15 & 0.9 & 71.1 & 0.15 & 0.9 \\
Precondition-checked simulation & 67.6 & 0.35 & 17.5 & 87.3 & 0.16 & 2.1 & 68.6 & 0.19 & 1.5 \\
Verdict only & 63.2 & 0.38 & 15.2 & 86.9 & 0.18 & 1.3 & 69.5 & 0.23 & 0.5 \\
\bottomrule
\end{tabular}
\caption{Listwise accuracy (\%), mean regret of the pointwise choice (extra steps), and share of pointwise choices that cannot be executed (\%).}
\label{tab:app_list}
\end{table*}

\begin{table}[t]
\centering
\footnotesize
\setlength{\tabcolsep}{2.5pt}
\begin{tabular}{@{}>{\raggedright\arraybackslash}p{3.1cm}rrrr@{}}
\toprule
Reason the action fails & $n$ & Llama & Qwen & Gemma \\
\midrule
Take an object not at hand & 658 & 77 & 18 & 6 \\
Move an object not held & 417 & 95 & 31 & 32 \\
Go to the current location & 209 & 100 & 22 & 98 \\
Open or close a receptacle elsewhere & 190 & 86 & 40 & 16 \\
Transform without object or appliance & 81 & 83 & 28 & 11 \\
Use a lamp elsewhere & 22 & 73 & 14 & 0 \\
Other & 13 & 46 & 38 & 23 \\
\bottomrule
\end{tabular}
\caption{Non-executable candidates by the reason they fail, and the share (\%) that each model simulates as executed.}
\label{tab:app_fail}
\end{table}

\begin{table}[t]
\centering
\small
\setlength{\tabcolsep}{3pt}
\begin{tabular}{lrrr}
\toprule
Predictor & Llama & Qwen & Gemma \\
\midrule
Correct without lookahead & 3.76$^{***}$ & 5.16$^{***}$ & 5.61$^{***}$ \\
Executability fidelity & 4.89$^{***}$ & 4.68$^{***}$ & 6.98$^{***}$ \\
Content F1 & 0.89$^{**}$ & -0.15 & -0.08 \\
Content defined & -0.48$^{*}$ & -0.15 & -0.26 \\
Simulation length (10 words) & 0.13 & 0.27 & 1.97$^{**}$ \\
\midrule
Corr.\ self-simulation, none & 0.944 & 0.965 & 0.952 \\
Corr.\ intent, none & 0.849 & 0.946 & 0.923 \\
Corr.\ self-simulation, intent & 0.827 & 0.938 & 0.905 \\
Corr.\ self-simulation, true & 0.898 & 0.940 & 0.917 \\
Corr.\ true, none & 0.929 & 0.938 & 0.908 \\
\bottomrule
\end{tabular}
\caption{Top: logistic regression of a correct self-simulation choice, with standard errors clustered by game ($^{*}p<0.05$, $^{**}p<0.01$, $^{***}p<0.001$). Bottom: correlation of candidate scores between notes after centering within each decision.}
\label{tab:app_reg}
\end{table}

Table~\ref{tab:app_list} reports the listwise scorer, which sees all four candidates with their notes and names the best one, together with the regret and the share of non-executable choices of the pointwise scorer. The listwise scorer is more sensitive to mismatched notes than the pointwise one, since it compares notes side by side. For Llama-8B and Qwen-27B it reproduces the conclusions of the main text: notes from the environment help, mismatched notes hurt, and self-simulation stays close to no lookahead. The listwise scorer of Gemma-31B is weak without notes (63.6\% against 85.3\% pointwise) and gains from any note that describes each candidate, including length-matched generic text (6.7 points) and intent (15.7 points), so for this model the listwise format measures the benefit of any per-candidate text and we rely on the pointwise scorer. Table~\ref{tab:app_fail} breaks down the non-executable candidates by the reason they fail. The weaker model reports almost every failed movement of an object and every redundant navigation as executed, which is the pattern of a simulation that narrates the intended effect of the action; Gemma-31B makes the same error for redundant navigation (98\%) while being accurate for most other failures. Table~\ref{tab:app_reg} gives the full regression of Section~\ref{sec:knows} and the correlations between candidate scores under different notes.

\section{Episodes}
\label{app:episodes}

Episodes use the 134 unseen games and stop after 40 steps or on success. At every step the agent samples eight completions of the proposer prompt at temperature 1 with at most 24 tokens, normalizes them to lower-case action strings, and keeps the four most frequent distinct actions. When only one distinct action remains, it is executed without scoring, so all lookahead conditions share the same proposer and differ only in the notes. Proposed actions are not filtered by admissibility, so the agent can propose and execute actions that cannot be executed, as LLM agents do in practice. The reactive baseline executes the greedy completion. The true-state condition replays the episode from the start for each candidate and restores the state afterwards. Success rates and their changes use 4{,}000 bootstrap resamples of games.

\begin{figure*}[t]
  \centering
  \includegraphics[width=\textwidth]{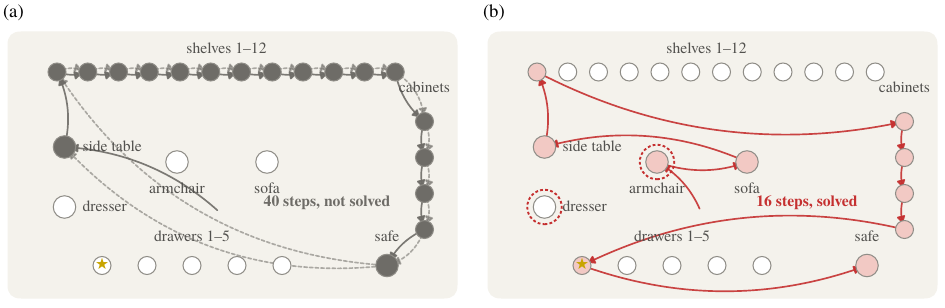}
  \caption{Self-simulation steers the search of Qwen-27B away from receptacles it has already seen. Routes in one unseen game (a) without lookahead and (b) with self-simulation, on a schematic layout of the room.}
  \label{fig:app_route}
\end{figure*}

The following excerpt illustrates the revisit pattern, and Figure~\ref{fig:app_route} draws both routes (filled circles mark visited receptacles, dashed arrows revisits, dashed rings the places simulated to hold a watch, and the star the drawer that holds it). In the unseen game ``put some watch on safe'', Qwen-27B without lookahead visits shelves 1 to 12 in order, opens two cabinets and the safe, and then visits shelves 1 to 12 again, ending without success. With self-simulation, its first candidates are ``go to armchair 1'' and ``go to dresser 1'', both simulated to show a watch, and after visiting the armchair, the sofa, the side table, one shelf, and four cabinets it opens drawer 1, finds the watch, and completes the task in 16 steps. Of the places it imagined, only the dresser held a watch, and the agent never went there; the simulations nevertheless steered the search toward plausible places for a watch and away from the shelves it had already searched without success.

\section{Artifacts and Compute}
\label{app:compute}

ALFWorld and TextWorld are released under the MIT license and are used here for research as intended. Llama-3.1-8B-Instruct is used under the Llama 3.1 Community License, Qwen3.8-27B under the Apache 2.0 license, and Gemma-4-31B-it under the Gemma Terms of Use, all for non-commercial research. The environment contains no personal data. All experiments ran on single NVIDIA H100 80GB GPUs with vLLM in bfloat16; the decision-level experiments took about 3.5 GPU hours and the episodes about 4 GPU hours in total. Decision points were built on CPUs in about 20 minutes.

\section{Prompts}
\label{app:prompts}
All roles share the context block shown below (task, initial observation, and up to the last ten steps). The proposer used in episodes additionally receives two expert demonstrations from the training split and the list of action templates.

\paragraph{Self-simulation.}
\begin{quote}\scriptsize\ttfamily
[system] You are a world model for a text-based household environment (ALFWorld). Given the task, the interaction so far, and a candidate action, predict the exact observation the environment will return immediately after the action is executed. Write it in the environment's own style, in one or two sentences. If the action cannot be executed in the current state, the environment returns "Nothing happens."\\{}
\\{}
Examples of observations in this environment:\\{}
- You arrive at cabinet 2. The cabinet 2 is closed.\\{}
- You arrive at countertop 1. On the countertop 1, you see a bread 1, a knife 2, and a mug 1.\\{}
- You open the fridge 1. The fridge 1 is open. In it, you see a egg 2, and a lettuce 1.\\{}
- You pick up the soapbar 1 from the toilet 1.\\{}
- You move the plate 1 to the shelf 2.\\{}
- You heat the potato 1 using the microwave 1.\\{}
- Nothing happens.\\{}
[user] Task: put a clean mug in coffeemachine.\\{}
\\{}
Initial observation: You are in the middle of a room. Looking quickly around you, you see a cabinet 1, a coffeemachine 1, a countertop 1, a sinkbasin 1, and a stoveburner 1.\\{}
\\{}
Interaction so far (oldest first):\\{}
> go to countertop 1\\{}
You arrive at countertop 1. On the countertop 1, you see a mug 1, and a spoon 2.\\{}
\\{}
Candidate action: take mug 1 from countertop 1\\{}
\\{}
Predicted observation:
\end{quote}

\paragraph{Intent only.}
\begin{quote}\scriptsize\ttfamily
[system] You are assisting a planner for an agent in a text-based household environment (ALFWorld). Given the task, the interaction so far, and a candidate action, state in one or two sentences what the agent would be trying to accomplish with this action and how it would contribute to the task. Do not predict what the environment will display and do not say whether the action will succeed.\\{}
[user] Task: put a clean mug in coffeemachine.\\{}
\\{}
Initial observation: You are in the middle of a room. Looking quickly around you, you see a cabinet 1, a coffeemachine 1, a countertop 1, a sinkbasin 1, and a stoveburner 1.\\{}
\\{}
Interaction so far (oldest first):\\{}
> go to countertop 1\\{}
You arrive at countertop 1. On the countertop 1, you see a mug 1, and a spoon 2.\\{}
\\{}
Candidate action: take mug 1 from countertop 1\\{}
\\{}
Purpose of this action:
\end{quote}

\paragraph{Precondition-checked simulation.}
\begin{quote}\scriptsize\ttfamily
[system] You are a world model for a text-based household environment (ALFWorld). Given the task, the interaction so far, and a candidate action, predict the exact observation the environment will return immediately after the action is executed. Write it in the environment's own style, in one or two sentences. If the action cannot be executed in the current state, the environment returns "Nothing happens."\\{}
\\{}
Examples of observations in this environment:\\{}
- You arrive at cabinet 2. The cabinet 2 is closed.\\{}
- You arrive at countertop 1. On the countertop 1, you see a bread 1, a knife 2, and a mug 1.\\{}
- You open the fridge 1. The fridge 1 is open. In it, you see a egg 2, and a lettuce 1.\\{}
- You pick up the soapbar 1 from the toilet 1.\\{}
- You move the plate 1 to the shelf 2.\\{}
- You heat the potato 1 using the microwave 1.\\{}
- Nothing happens.\\{}
\\{}
Before predicting, check whether the action's preconditions hold in the current state. For example, the agent must already be at a receptacle to open it, take an object from it, or put an object into it; it must be holding an object to move, heat, cool or clean it; the object must have been seen at the agent's current location to be taken; and the agent cannot go to the location it is already at. Answer in exactly this format:\\{}
Preconditions: <one short sentence>\\{}
Observation: <predicted observation>\\{}
[user] Task: put a clean mug in coffeemachine.\\{}
\\{}
Initial observation: You are in the middle of a room. Looking quickly around you, you see a cabinet 1, a coffeemachine 1, a countertop 1, a sinkbasin 1, and a stoveburner 1.\\{}
\\{}
Interaction so far (oldest first):\\{}
> go to countertop 1\\{}
You arrive at countertop 1. On the countertop 1, you see a mug 1, and a spoon 2.\\{}
\\{}
Candidate action: take mug 1 from countertop 1
\end{quote}

\paragraph{Pointwise scorer.}
\begin{quote}\scriptsize\ttfamily
[system] You are evaluating candidate next actions for an agent in a text-based household environment (ALFWorld). The agent must complete its task in as few steps as possible.\\{}
[user] Task: put a clean mug in coffeemachine.\\{}
\\{}
Initial observation: You are in the middle of a room. Looking quickly around you, you see a cabinet 1, a coffeemachine 1, a countertop 1, a sinkbasin 1, and a stoveburner 1.\\{}
\\{}
Interaction so far (oldest first):\\{}
> go to countertop 1\\{}
You arrive at countertop 1. On the countertop 1, you see a mug 1, and a spoon 2.\\{}
\\{}
Candidate action: take mug 1 from countertop 1\\{}
Lookahead note: You pick up the mug 1 from the countertop 1.\\{}
\\{}
How promising is this candidate as the next action for completing the task efficiently? Rate it from 1 (useless or counterproductive) to 9 (clearly the best next step). Answer with a single digit.
\end{quote}

\paragraph{Listwise scorer.}
\begin{quote}\scriptsize\ttfamily
[system] You are evaluating candidate next actions for an agent in a text-based household environment (ALFWorld). The agent must complete its task in as few steps as possible.\\{}
[user] Task: put a clean mug in coffeemachine.\\{}
\\{}
Initial observation: You are in the middle of a room. Looking quickly around you, you see a cabinet 1, a coffeemachine 1, a countertop 1, a sinkbasin 1, and a stoveburner 1.\\{}
\\{}
Interaction so far (oldest first):\\{}
> go to countertop 1\\{}
You arrive at countertop 1. On the countertop 1, you see a mug 1, and a spoon 2.\\{}
\\{}
Candidate actions:\\{}
(A) take mug 1 from countertop 1\\{}
    Lookahead note: You pick up the mug 1 from the countertop 1.\\{}
(B) go to sinkbasin 1\\{}
    Lookahead note: You arrive at sinkbasin 1. On the sinkbasin 1, you see nothing.\\{}
\\{}
Which candidate is the best next action for completing the task efficiently? Answer with a single letter.
\end{quote}

\end{document}